\documentclass{article}
\usepackage{pslatex}
\usepackage{graphics}
\usepackage{rotating}
\usepackage[table]{xcolor}
\usepackage{caption}
\usepackage{hyperref}
\usepackage{amsthm}
\usepackage{amsmath}
\usepackage{enumitem}
\usepackage{amssymb}
\theoremstyle{plain}
\usepackage{tabularx}
\usepackage{multirow}

\begin{document}

\title{Sherlock Holmes Doesn't Play Dice:\\
The mathematics of uncertain reasoning\\when something may happen,\\that one is not even able to figure out}

\author{Guido Fioretti\\University of Bologna\\
Contact address: guido.fioretti@unibo.it}
\maketitle

\begin{abstract}
While Evidence Theory (also known as Dempster-Shafer Theory, or Belief Functions Theory) is being increasingly used in data fusion, its potentialities in the Social and Life Sciences are often obscured by lack of awareness of its distinctive features. In particular, with this paper I stress that an extended version of Evidence Theory can express the uncertainty deriving from the fear that events may materialize, that one is not even able to figure out. By contrast, Probability Theory must limit itself to the possibilities that a decision-maker is currently  envisaging.

I compare this extended version of Evidence Theory to sophisticated extensions of Probability Theory, such as imprecise and sub-additive probabilities, as well as unconventional versions of Information Theory that are employed in data fusion and transmission of cultural information. A further extension to multi-agent interaction is outlined.
\end{abstract}


\newpage

\section{Introduction}     \label{sec:intro}

Sometimes, unexpected and novel events upset the network of causal relations on which we base our decisions. A global pandemic in the XXI century, a local conflict that might degenerate into World War III, as well the 2008 financial crisis are global and well-known examples of destructive events that nobody had  conceived before they actually happened. One consequence of such events is that they may suggest a  simple but fundamental question: ``Now, what else?'' 

Such  questions do not follow automatically from empirical evidence, but rather depend on existing mental models and whether we allow them to be questioned by that evidence.  Dr. Watson typically sticks to the most obvious interpretation of facts, whereas Sherlock Holmes allows apparently negligible cues to upset the received wisdom. Does Sherlock Holmes know from the very first page who is guilty? Obviously, not. But he knows that, whatever the truth, it will be different from anything he can currently figure out. Therefore, he starts searching for an alterative explanation.

Probability Theory (PT) cannot express the doubt that  novel and potentially disruptive possibilities may materialize, precisely because one of its assumptions is that an exhaustive set of   possibilities is  given. Nevertheless, this sort of non-probabilistc uncertainty has been hotly debated in economics because of its impact on  investment decisions \cite{kay-king-20}.

This sort of uncertainty, which has been variously qualified  as ``Keynesian,'' ``fundamental,'' ``true,'' ``epistemic,'' ``ontological'' and more recently ``radical'' uncertainty \cite{runde-90EP} \cite{davidson-91JEP} \cite{dunn-01JPKE} \cite{dequech-04CJE} \cite{lane-maxfield-05JEE} \cite{kay-king-20} should be clearly distinguished from the uncertainty deriving from lack of information on given possibilities, such as the paucity of the sample on which probabilities are measured. Small sample size, unfair dice and  unique events pose serious problems to the assessment of reliable probabilities, but they still concern a set of known events. With a possibly awkward but dense expression the literature traces a clear distinction between ``known unknowns'' (unknown probabilities of known possibilities) and ``unknown unknowns'' (unknown probabilities of unknown possibilities) \cite{rumsfeld-11} \cite{feduzi-runde-14OBHDP} \cite{faulkner-feduzi-runde-17CJE}. 

While the problem of known unknowns is interesting in itself, I rather focus on the more challenging problem posed by unknown unknowns. Specifically, the reason for writing this essay is that Evidence Theory (ET) \cite{shafer-76} and its subsequent improvements provide a framework to deal with unknown unknowns by not assuming that the complementation operator is necessarily applied to the possibility set. Without complementation, the easy but illusory solution of defining an all-encompassing residual event is not available. In other words, the disturbing clues that stimulate Sherlock Holmes's investigations cannot be moved under the carpet.

The canonical setting of ET is a judge listening to testimonies or a detective looking for clues, rather than a gambler playing dice \cite{shafer-81S} \cite{shafer-tversky-85CS}. This is critical, because while it is natural for a gambler to reason in terms of a given set of outcomes,  a detective must be open to unexpected denouements. While a substantial portion of the literature  has limited itself to the purely technical aspects of ET, I stress its distinctive paradigm and its logical consequences. In particular, I aim at connecting with one another the awareness of the importance of radical uncertainty developed by social scientists with the ability of certain extended versions of ET to deal with it.

The rest of this paper is organized as follows. The ensuing section \S~\ref{sec:ET} illustrates the basics of ET with respect to both known and unknown unknowns. Contrary to most introductions to ET, I emphasize its  degree of freedom concerning hypotheses formulation. Subsequently, \S~\ref{sec:ET-PT-IT} frames ET with respect to PT and Information Theory (IT), respectively. In particular, \S~\ref{subsec:ET-PT} illustrates the ability of ET to estimate known unknowns  whereas \S~\ref{subsec:ET-IT} discusses the usefulness of  ET for non-trivial problems of   information transmission.  Section\S~\ref{sec:CSN} explores the possibility of using ET in interpersonal decision-making.   Finally, section \S~\ref{sec:conclusions} concludes with prospects for future  developments and applications.

\section{Radical Uncertainty Within Evidence Theory}     \label{sec:ET}

This brief introduction to ET aims at highlighting its ability to express radical uncertainty. The ensuing account merges  Shafer's original ET \cite{shafer-76} with insights from  Smets's Transferable Belief Model (TBM) \cite{smets-92IJIS}. In particular, the possibility to express radical uncertainty is a nice contribution of TBM that I shall extend insofar it concerns its consequences for representing decision-making (see \S~\ref{app:merge} for a detailed account of contributions and interpretations).

In ET, the possibility set is called \emph{frame of discernment} (FoD) in order to stress that it represents what possibilities a decision-maker is envisaging at a certain point in time. Let  $\Theta$ denote a FoD that entails possibilities $A_1,$ $ A_2, \ldots$ $A_N$. These possibilities are supported by masses of empirical evidence $m(A_1),$ $ m(A_2), \ldots$ $m(A_N)$.

Notably, possibilities $A_i$ are not necessarily disjoint sets. Testimonies support one another insofar they correspond to (partially) overlapping sets, whereas they contradict one another insofar they are disjoint. In general, $\forall i,j$ it is possible that $A_i \cap A_j \neq \emptyset$. Formally, PT can be obtained as a special case of ET when all possibilities are singletons (see \S~\ref{subsec:ET-PT} for details).

Since in the FoD possibilities are represented as sets that may intersect with one another, $m(A_i) + m(A_j)$ is generally not equivalent to $m(A_i \cup A_j)$. Thus, albeit normalization to 1 is possible and generally carried out,  it does not imply that a given amount is distributed among mutually exclusive possibilities as in PT.

One other consequence of representing possibilities as sets instead of singletons is that while ET  is able to deal with  ``known unknowns'' by means of smaller masses $m$ pretty much as PT does by assuming  sub-additive probabilities,  ET can also approach the problem of  ``known unknowns'' by  tuning  overlaps between possibilities. Consider, for instance, the novel natural catastrophes that are being caused by climate change. Since the sample is too small for a reliable estimation of probabilities, PT  resorts to sub-additive probabilities. Likewise, ET can resort to assigning small values to masses $m$. However, ET can also assess the  differences of climate change -induced catastrophes with the past ones,  translating these  differences into the contours of the corresponding possibilities in the FoD (see \S~\ref{subsec:ET-PT} for details).

Evidence Theory comes in flavours, where frame of discernment $\Theta$ may or may not be coupled to the complementation operator to form a  $\sigma$-algebra. Henceforth, I shall assume that $\Theta$ is not a $\sigma$-algebra in order to exclude the possibility of encapsulating unknown unknowns into an all-encompassing residual event.

Radical uncertainty (due to unknown unknowns)  originates from evidence that contradicts established relations of causes and effects, with novel possibilities eventually entering the FoD. In other words, once novel and unthinkable things have been observed, one expects other unknown unknowns to appear in the future. By leveraging on empirical investigations and theoretical considerations on abductive logic \cite{locke-goldenbiddle-feldman-08OS} \cite{altmann-16II} \cite{saetre-vandeven-21AMR}, I take the amount of conflicting evidence as an appropriate measure of radical uncertainty. In other words, the more the conflicting evidence, the stronger the current causal maps are questioned, hence the stronger the fear that they may be finally upset.

Within ET, the Transferable Belief Model (TBM) \cite{smets-88} \cite{smets-92IJIS} \cite{smets-92XL} assumes that conflicting evidence translates into $m(\emptyset) > 0$. The rationale of this assumption  is that conflicting evidence, by suggesting that something may happen, that is currently not in the FoD, moves some mass $m$ towards possibilities that cannot yet be defined.

Independently of radical uncertainty,  ET also allows to assign a positive mass to the FoD as a whole. This mass is neither committed to the possibilities that are being envisaged, nor to the void set which represents the fear that something else may happen. An $m(\Theta) >0$ represents suspension of judgement, non-assigned belief that in the course of investigations the judge or detective may assign or withdraw from  specific possibilities, or the void set. In the final denouement of detective stories, both $m(\emptyset)$ and $m(\Theta)$ finally shrink down to zero.

Let us suppose that  possibilities $A_1,$ $ A_2, \ldots$ $A_N$ are being envisaged,  supported by masses of empirical evidence $m(A_1),$ $ m(A_2), \ldots$ $m(A_N)$, respectively. Let us suppose that a state of mind expressing the fear of unknown unknowns is there, which translates into  $m(\emptyset)>0$. Let us suppose that  evidence is sufficiently scant to suggest restraint, which translates into   $m(\Theta)>0$. Though not essential to the theory, masses $m(.)$ can be normalized in order to obtain that:

\begin{equation}
  \sum_{i=1}^{N} \: m_i(A_i) \; + \; m_A(\emptyset) \; + \; m_A(\Theta) \; = \; 1         \label{eq:m}
\end{equation}

Notably, since in ET possibilities $A_i$ are not necessarily disjoint sets, eq.~\ref{eq:m} does not amount to distributing a given mass among distinct possibilities. This  normalization adjusts the evidence supporting partially overlapping sets, net of judgement restraint represented by $m(\Theta)$ and the fear of unknown unknowns represented by $m(\emptyset)$.

Let us assume that evidence   $A= \{m(A_1),$ $ m(A_2), \ldots$ $m(A_{N_A}),$ $m_A(\emptyset),$ $m_A(\Theta) \}$ is available when a new body of evidence arrives (e.g., a new testimony, new cues, etc.). Let   $B = \{ m(B_1),$ $ m(B_2), \ldots$  $m(B_{N_B}),$ $m_B(\emptyset),$ $ m_B(\Theta) \}$ be this new body of evidence. Just like the sets entailed in one single body of evidence are not necessarily disjoint, $\forall i,j$ it may either be $A_i \subseteq B_j$, or $A_i \supseteq B_j$, or $Ai \cap B_j \neq \emptyset$, or $A_i \cap B_j = \emptyset$.

The judge, or detective, must evaluates which items in these two bodies of evidence are coherent with one another while weighing them against contradictory items. In the closed world assumption (no unknown unknowns, and therefore $m_A(\emptyset) =$ $ m_B(\emptyset) =$ $m_C(\emptyset)= 0$), Dempster-Shafer's combination rule \cite{dempster-68JRSSB} \cite{shafer-76} yields the components of a new body of evidence $C$ that unites  $A$ and $B$. Note that intersections with $\Theta$ enter the computation. 

\begin{equation}
  m(C_k) \; = \; \frac{\sum_{X_i \cap Y_j = C_k} \; m_A(X_i) \, m_B(Y_j)}{1 - \: \sum_{X_i \cap Y_j = \emptyset} \; m_A(X_i) \, m_B(Y_j)}  \label{eq:dempster-shafer}
\end{equation}
where $X_i \in \{ A_i \, \forall i, \; \Theta  \}$,  $Y_j \in \{ B_j \, \forall j, \; \Theta  \}$, and  where the $C_k$s are defined by all possible intersections of the $X_i$s with the $Y_j$s.

The numerator of eq.~\ref{eq:dempster-shafer} measures the extent to which the two bodies of evidence support $C_k$, whereas the denominator measures the extent to which they are not contradictory with one another. In the simplest, 1-dimensional case \cite{shafer-86IJIS},  information is conveyed through a series of $n$ testimonies of reliability $m$ each, yielding a combined reliability $m^n$. By contrast, the combined reliability of $n$ independent parallel testimonies is $1- (1-m)^n$. Thus, in eq.~\ref{eq:dempster-shafer} the numerator expresses the logic of serial testimonies whereas the denominator expresses the logic of parallel testimonies.

Dempster-Shafer’s combination rule~\ref{eq:dempster-shafer} can be iterated to combine any number of evidence bodies. Its outcome is independent of the order in which they are combined.

Rule~\ref{eq:dempster-shafer} has been found to yield unsatisfactory results for closed-world problems that are framed in  terms of mutually exclusive possibilities. Several solutions have been proposed, including alternative combination rules  (see \S~\ref{app:dezert} for an illustration of alternatives). However,  mutually exclusive possibilities are typical of gamblers playing games of chance, rather than judges listening to testimonies or detectives evaluating cues. Possibly, those are not the sort of decision problems to which ET should be applied \cite{haenni-05WP} \cite{boivin-22WP}.

The case of an open world where unknown unknowns are possible (either $m_A(\emptyset)>0$, or $m_B(\emptyset) > 0$, or both) is substantially more complex. Conflict can neither be ignored nor redistributed. In open worlds,  conflict is an indicator that mental models are failing, causal maps are not providing orientation, and some cognitive re-arrangement is in order. Novel possibilities are likely to emerge, that are currently  impossible to conceive.

In an open world, TBM  applies. TBM makes use of the numerator of eq.~\ref{eq:dempster-shafer} while extending it to $m(\emptyset)$ \cite{smets-88} \cite{smets-92XL}. The idea is that since $m(\emptyset)>0$ expresses a degree of belief concerning possibilities that are currently not within $\Theta$, it should not be redistributed among those that are:

\begin{equation}
  m(C_k) \; = \; \sum_{X_i \cap Y_j = C_k} \; m_A(X_i) \, m_B(Y_j)  \label{eq:smets}
\end{equation}
where $X_i \in \{A_i \, \forall i, \: \emptyset, \: \Theta\}$,  $Y_j \in \{B_j \, \forall j, \: \emptyset, \: \Theta\}$ and the $C_k$s are defined by all possible intersections of the $X_i$s with the $Y_j$s.

The outcome of Smets' combination rule must be  normalized by means of eq.~\ref{eq:m}. While normalization is optional for eq.~\ref{eq:dempster-shafer}, it is compulsory for eq.~\ref{eq:smets}.

Let us  suppose that, once all available bodies of evidence have been combined, the  judge or detective  formulates a hypothesis $\mathcal{H}$. This hypothesis is a subset of $\Theta$ but, unlike the $A_i$s, it does not represent empirical evidence  but rather a mental construct.

The belief that the judge or detective can reasonably attach to  $\mathcal{H}$ is given by the amount of evidence supporting it. Assuming a body of evidence $C = \{ m(C_1),$ $ m(C_2), \ldots$   $m(C_{N_C}),$ $m_C(\emptyset),$ $m_C(\Theta) \}$, and in accord with TBM \cite{smets-07IF},  the following \emph{Belief Function} expresses the belief in $\mathcal{H}$ supported by $C$:

\begin{equation}
  Bel(\mathcal{H}) \; = 
  \left\{
  \begin{array}{lll}
  \sum_{C_k \subset \mathcal{H}} \; m(C_k) & \mathrm{if} & \mathcal{H} \subset \Theta, \; \mathcal{H} \neq \emptyset \\    \label{eq:belief}
  m_C(\emptyset) & \mathrm{if} & \mathcal{H} \equiv \emptyset \\ 
  m_C(\Theta) & \mathrm{if} & \mathcal{H} \equiv \Theta 
  \end{array}
  \right.      
\end{equation}
where the second and third lines are different from the  formulation for closed worlds \cite{shafer-76},  which assumed  $Bel(\emptyset) = 0$ and $Bel(\Theta) = 1$.

While belief in $\mathcal{H}$ is only supported by the evidence  bearing specifically on  $\mathcal{H}$, it may be desirable to include also the evidence that partially supports it. In particular, as in TBM \cite{smets-07IF} I introduce the following  \emph{Plausibility Function}:

\begin{equation}
  Pl(\mathcal{H}) \; = 
  \left\{
  \begin{array}{lll}
  \sum_{C_k \cap \mathcal{H} \neq \emptyset} \; m(C_k) & \mathrm{if} & \mathcal{H} \subset \Theta, \; \mathcal{H} \neq \emptyset \\    \label{eq:plausibility}
  m_C(\emptyset) & \mathrm{if} & \mathcal{H} \equiv \emptyset \\ 
  m_C(\Theta) & \mathrm{if} & \mathcal{H} \equiv \Theta 
  \end{array}
  \right.    
\end{equation}
where the second and third lines are different from the formulation for closed worlds \cite{shafer-76},  which  assumed $Pl(\emptyset) = 0$ and $Pl(\Theta) = 1$.

Obviously, $Bel(\mathcal{H}) \leq Pl(\mathcal{H})$. In many applications, belief is more important than plausibility.

Note also that while the frame of discernment is not allowed to generate novel possibilities by complementation (no residual events), the judge or detective  can make use of any sort of operator in order to formulate hypotheses. For instance, one can assume that hypotheses are formulated in a subset $\Omega \subset 
 \Theta$ which is a   $\sigma$-algebra  \cite{kohlas-monney-08XXVI}. Thus, $\forall \mathcal{H} \neq \emptyset$ and  $\forall \mathcal{H} \neq \Theta$ it is $Bel(\mathcal{H}) + Bel(\overline{\mathcal{H}}) \leq 1$ and $Pl(\mathcal{H}) + Pl(\overline{\mathcal{H}}) \geq 1$. 

In general, decision-makers may formulate several alternative hypotheses, which they may wish to compare to one another given the available evidence. For instance, hypotheses $\mathcal{H}_1$ and  $\mathcal{H}_2$ might be compared by evaluating either $Bel(\mathcal{H}_1) \lessgtr Bel(\mathcal{H}_2)$, or $Pl(\mathcal{H}_1) \lessgtr Pl(\mathcal{H}_2)$.

In general, hypotheses change with time.
The alternative hypotheses that are being entertained can either change out of some behavioural algorithm simulating human reasoning, or actual participation of a  human being in subsequent interactions with an expert system, or they may be simply suggested by subsequent iterations of eq.~\ref{eq:dempster-shafer} or~\ref{eq:smets}, in which case $\mathcal{H}_k \equiv C_k$. Or, some combination of the above cases. Note that ET does not impose any constraint on the process of hypotheses generation.

Hypotheses generation can be just as trivial as those generated by Dr. Watson, or as creative and surprising as those conceived by Sherlock Holmes.  Since several experiments have established that conflicting evidence impairs decision-making  \cite{tversky-shafir-92PSdeferred} \cite{gluth-rieskamp-buchel-13PLOSCB} \cite{yin-devreede-steele-devreede-19SS} \cite{yin-devreede-steele-devreede-21ICIS}, I submit that creative hypotheses generation is triggered when  $m(\emptyset)$ goes beyond some threshold. Estimations of this threshold are only available for specific experiments \cite{gluth-rieskamp-buchel-13PLOSCB}, and are likely to be moderated by factors that are still unknown. Heterogeity certainly exists, with Sherlock Holmes characterized by a much lower threshold than Dr. Watson.

Hypothesis generation implies tightening or coarsening the FoD. While this aspect is generally neglected in the literature on ET, Glenn Shafer made a few illuminating remarks in this respect:

\begin{quotation}
Like any creative act, the act of constructing a frame of discernment  does not lend itself to thorough analysis. But we can pick out two considerations that influence it: (1) we want our evidence to interact in an interesting way, and (2) we do not want it to exhibit too much internal conflict.

Two items of evidence can always be said to interact, but they interact in an interesting way only if they jointly support a proposition more interesting than the propositions supported by either alone. (...) Since it depends on what we are interested in, any judgment as to whether our frame is successful in making our evidence interact in an interesting way is a subjective one. But since interesting interactions can always be destroyed by loosening relevant assumptions and thus enlarging our frame, it is clear that our desire for interesting interaction will incline us towards abridging or tightening our frame.

Our desire to avoid excessive internal conflict in our evidence will have precisely the opposite effect: it will incline us towards enlarging or loosening our frame. For internal conflict is itself a form of interaction --- the most extreme form of it. And it too tends to increase as the frame  is tightened, decrease as it is loosened.

\end{quotation}
Glenn Shafer \cite{shafer-76}, Ch. XII.

\bigskip

While the early versions of TBM  proceeded to a ``pignistic transformation'' to probabilities whenever $m(\emptyset)>0$  \cite{smets-92IJIS} \cite{smets-kennes-94AI}, more recent applications are capable of attaching lower reliability to certain bodies of evidence  \cite{rakar-juricic-02JPR} \cite{ramasso-panagiotakis-rombaut-pellerin-tziritas-08ELCVIA} \cite{ioannou-louvieris-clewley-09IEEE-A} or discount correlations between different bodies of evidence  \cite{benyaghlane-denoeux-mellouli-01XXXII} \cite{haduong-08IJAR}. However, ET's philosophy is that, just like Sherlock Holmes looks for  details that finally overthrow Dr. Watson's interpretation, the frame of discernment should be tightened and coarsened until the sources are either sufficiently detailed to be reliable and uncorrelated with one another, or discarded altogether \cite{shafer-16IJAR-hist}.

ET is designed for iteratively zooming the frame of discernment until the judge, or detective, arrives at an interesting, non-trivial representation that expresses little or no contradiction \cite{gordon-shortliffe-90VII-III}. One  research direction, not yet explored, could consist of using ET in human-machine interaction.

\section{Evidence, Probability, and Information Theory}   \label{sec:ET-PT-IT}

This section illustrates PT and IT from the point of view of ET. Formally, ET understands PT and IT as special cases that obtain when sets $A_i$ are singletons  $\{A_i\}$ representing possibilities that can either be  distinct or coincide, but never intersect. It is also necessary to assume $m(\emptyset)=0$ whereas a sort of $m(\Theta)>0$ can be contemplated by certain versions of PT (sub-additive probabilities). With these assumptions, the mathematics of PT and IT becomes a subset of ET. However, the conceptual difference between a gambler and a judge, or detective, is bound to stay.

\subsection{Evidence Theory and Probability Theory}    \label{subsec:ET-PT}

Technical and practical differences between ET and PT become apparent when Dempster-Shafer's combination rule is compared to Bayes' rule \cite{challa-koks-04S} \cite{dezert-wang-tchamova-dezert-15XIX} \cite{tchamova-dezert-15XXI}. While there exist several accounts of  specific cases where Dempster-Shafer combination rule~\ref{eq:dempster-shafer} can be interpreted within PT \cite{smets-94I}, I rather explore how Bayes's rule can be understood within ET.

In its basic version, PT implies --- among else --- the following assumptions:

\begin{enumerate}[label = (\roman*)]
  \item All possibilities are singletons, in which case $\forall \{A_i\}$ and $\forall \{B_j\}$ it is either $\{A_i\} \cap \{B_j\} \equiv$ $ \{A_i\} \equiv$ $ \{B_j\}$ or $\{A_i\} \cap \{B_j\} = \emptyset$. In other words, possibilities are not sufficiently nuanced to enable partial overlap. Since it is not possible to generate possibilities beyond those that are included in the incoming bodies of evidence, no novel $C_k$ can be generated by eq.~\ref{eq:dempster-shafer}.   \label{cond:singletons}

  \item Although novel possibilities can present themselves, no belief can be allocated to the fear that this may happen. Thus, $m(\emptyset)=0$. Moreover, insufficient sample size is countered by the Principle of Sufficient Reason. Thus,  $m(\Theta) = 0$ and the bodies of evidence to be combined take the form    $p(\{A_1\}),$ $ p(\{A_2\}), \ldots$ $p(\{A_{N_A}\})$ and  $p(\{B_1\}),$ $ p(\{B_2\}), \ldots$ $p(\{B_{N_B}\})$, respectively, where $N_A, N_B \in \mathbb{N}$. Probabilities $p$  are subject to the usual constraints $\sum_{i} p(\{A_i\}) = 1$ and $\sum_{j} p(\{B_j\}) = 1$.   \label{cond:radicalUncertainty}
\end{enumerate}

In this special case, Dempster-Shafer combination rule (eq~\ref{eq:dempster-shafer}) boils down to Bayes' Theorem (see \S~\ref{app:bayes} for details). However, PT has been greatly extended beyond  assumptions~\ref{cond:singletons} and~\ref{cond:radicalUncertainty}. In particular, \emph{imprecise probabilities} can be defined over an interval $[p_*, p^*]$ where $p_*$  and $p^*$ are called \emph{lower probability} and   \emph{upper probability}, respectively. Empirical measurement is expected to elicit that  $p \in$ $[p_*, p^*]$ rather than assessing the exact value of $p$.

Imprecise probabilities are not additive, for $\sum_i p_*(\{A_i\}) \leq 1 $ and $\sum_i p^*(\{A_i\}) \geq 1 $. However, $p^*(\{A_i\}) = 1 - p_*(\{\overline{A}_i\}), \forall i$. With imprecise probabilities, assumption~\ref{cond:singletons} does not change, whereas~\ref{cond:radicalUncertainty} does:

\begin{enumerate}[label = (\roman*)']
  \item $\equiv \;\;$  \ref{cond:singletons}   \label{cond:singletons1}

  \item Although novel possibilities can present themselves, no belief can be allocated to the fear that this may happen. Thus, $m(\emptyset)>0$. However, it is generally $m(\Theta) \geq 0$, with strict inequality if at least one probability is lower than $p^*$. The bodies of evidence to be combined take the form  $\{ [p_*(\{A_1\}),$ $p^*(\{A_1\})],$ $ [p_*(\{A_2\}),$ $ p^*(\{A_2\})]  \ldots$ $[p_*(\{A_{N_A}\}),$ $p^*(\{A_{N_A}\})] \}$ and  $\{ [p_*(\{B_1\}),$ $ p^*(\{B_1\})],$   $[p_*(\{B_2\}),$ $p^*(\{B_2\})], \ldots$  $[p_*(\{B_{N_B}\}),$ $ p^*(\{B_{N_B}\})] \}$, respectively, where $N_A, N_B \in \mathbb{N}$.    \label{cond:radicalUncertainty1}
\end{enumerate}

Imprecise  probabilities can be used to combine probabilistic uncertainty with the uncertainty deriving from relying on too small a sample --- the known unknowns.
Suppose, for instance, that you are playing for the first time with a die that you suspect may not  be fair.  Lack of information may prudentially suggest  $p \in [1/7, \, 1/5]$  rather than $p = 1/6$. Later on, by throwing the die again and again this interval shrinks down towards the true, precise probability.

When imprecise probabilities are employed in order to deal with  known unknowns, upper probabilities are sometimes neglected. The remaining lower probabilities are eventually called \emph{sub-additive probabilities}, to which the standard probability calculus  applies \cite{gilboa-87JME} \cite{sarin-wakker-92E}. In particular, a body of evidence   $\{ p_*(\{A_1\}),$ $ p_*(\{A_2\}),$ $\ldots$ $p_*(\{A_{N_A}\}) \}$ can be conditioned on  $\{ p_*(\{B_1\}),$ $p_*(\{B_2\}),$ $ \ldots$ $p_*(\{B_{N_B}\}) \}$ by means of Bayes's rule \cite{jaffray-92IEEE-SMC}.

More in general, imprecise probabilities on singletons can be handled just like precise probabilities on partially overlapping sets \cite{walley-91} \cite{ferson-kreinovich-ginzburg-myers-sentz-03WP}. In order to grasp the rationale for this transformation, suppose that you are dealing with an unfair die where face 1 shows up more often than $1/6$ because some lead has been injected just below face 6. Thus, faces 2, 3, 4 and 5 show up less often and face 6 least often. You can understand it as if a small portion of the events ``face 2'' to ``face 5,'' and a large portion of the event ``face 6'' are turned into the event ``face 1.'' For instance, you should have observed face 2, but you observe face 1 in fact.

\begin{figure}
\center
\fbox{\resizebox*{0.8\textwidth}{!}{\includegraphics{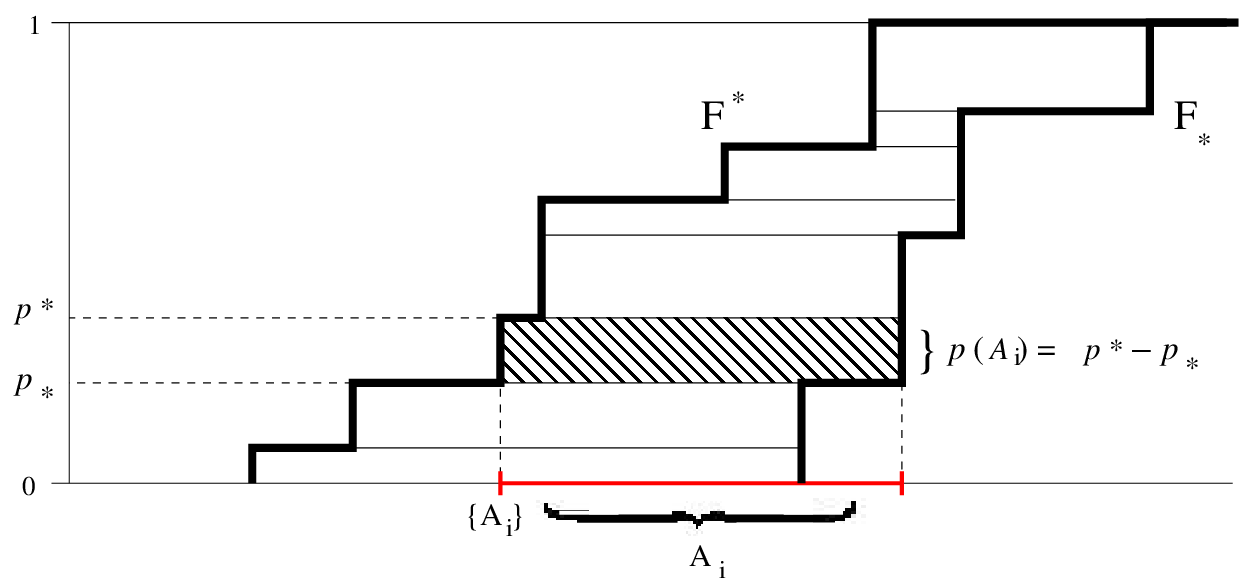}}}
\caption{Transformation of imprecise probabilities defined on singletons into single-valued probabilities defined on intervals. Step-wise cumulative lower probability function $F_*$ and cumulative upper probability function $F^*$ identify intervals $A_i$ with probability $p(A_i) = p^* - p_*$. Notably, for any $(i, i\,')$ it may happen that $A_i \cap A_{i\:'} \neq \emptyset$.} \label{fig:p-box}
\end{figure}

Figure~\ref{fig:p-box} illustrates this transformation for 1-dimensional sets. The lower and upper cumulative functions $F_*$ and $F^*$ delimit a probability interval $[p_*, p^*]$ for the singleton  $\{A_i\}$. This is the standard format for imprecise probabilities. However,  it can also be expressed in terms of a possibility set $A_i$ and a single-valued probability $p(A_i) = p^* - p_*$.

The transformation illustrated in Figure~\ref{fig:p-box} has practical significance.
Consider  insurance companies facing the problem of evaluating the cost of adverse events without reliable samples, which is  the first instance of known unknowns ever identified in economics \cite{knight-21}. For instance, climate change  favours wildfires, hence the probabilities that had been measured decades ago no longer apply. This uncertainty concerns a known possibility, namely wildfires, but  its probability is unknown --- it is a known unknown, indeed. The theory of imprecise probabilities suggests to use probability intervals, which is theoretically sound but offers no guidance as to how the extremes of these intervals might be computed. However, the transformation illustrated in Figure~\ref{fig:p-box} suggests that one may rather attempt to look into technical differences between the climate-change-induced wildfires with respect to the purely natural ones, for instance in terms of the length of the dry season, firefighters' equipment,  the strength and direction of winds in specific areas, or else. These features correspond to a set of possibilities that partially overlaps with that of purely natural wildfires by an extent which a technical evaluation can assess.

Note that with the transformation illustrated in Figure~\ref{fig:p-box}  we obtain the framework of ET, which is based on sets $A_i$ rather than singletons $\{A_i\}$. This transformation is  not always one-to-one because of a few special cases when singletons appear along with intervals, though  it is one-to-one in most practical applications \cite{walley-91} \cite{ferson-kreinovich-ginzburg-myers-sentz-03WP}.

The duality of singleton-based imprecise probabilities and set-based single-valued probabilities suggests to re-formulate assumptions~\ref{cond:singletons1} and~\ref{cond:radicalUncertainty1} as follows:

\begin{enumerate}[label = (\roman*)'']
  \item Possibilities are generally represented by sets $A_1,$ $A_2 \ldots$ $A_{N_A}$, which may intersect with one another. Thus, novel possibilities $C_k$ can be generated by eq.~\ref{eq:dempster-shafer}.   \label{cond:singletons2}

  \item Although novel possibilities can present themselves, no belief can be allocated to the fear that this may happen. Thus, $m(\emptyset)=0$. However, $m(\Theta) \geq 0$ and the bodies of evidence to be combined take the form    $p(A_1),$ $ p(A_2),$ $ \ldots$ $p(A_{N_A}),$ $m_A(\Theta)$ and  $p(B_1),$ $ p(B_2),$ $ \ldots$ $p(B_{N_B}),$ $m_B(\Theta)$, respectively, where $N_A, N_B \in \mathbb{N}$.  \label{cond:radicalUncertainty2}
\end{enumerate}

With assumptions~\ref{cond:singletons2} and~\ref{cond:radicalUncertainty2} we are still within PT, but bodies of evidence must be combined by means of Dempster-Shafer rule (eq~\ref{eq:dempster-shafer}) instead of Bayes's Theorem. The main differences with ET are that (a) Probabilities $p$ appear instead of  masses $m$, and (b) The possibility that $m(\emptyset) > 0$ is ignored.

One remarkable conclusion is that  Dempster-Shafer's combination rule, as well as belief and plausibility functions defined on the $C_k$ induced by eq~\ref{eq:dempster-shafer}, are well within (an extended version of) PT. Indeed, Arthur Dempster moved initially from imprecise probabilities  when he proposed eq.~\ref{eq:dempster-shafer} \cite{dempster-68JRSSB}. The framework of a judge or detective looking for cues  makes the difference between ET and PT, not the maths.

\subsection{Evidence Theory and Information Theory}    \label{subsec:ET-IT}

IT \cite{shannon-weaver-49} assumes that a  source emits characters drawn from a known alphabet $ A= \{ \{A_1\},$ $ \{A_2\},$ $ \ldots \{A_N\}  \}$. These characters must travel through a noisy channel in order to be communicated to a receiver who is aware that the characters have been drawn from $A$. Noise is able to alter characters. Thus, in order to minimize errors each character $\{A_i\}$ is coded into a set of characters $A_i$, with $\parallel A_i \parallel>1$ where $\parallel A_i \parallel$ is the length of the sequence of characters into which each original character is coded. Since  noise is unlikely to alter sufficiently many characters of  $A_i$ to make it unrecognizable, the receiver is most often able to reconstruct the original character. The greater $\parallel A_i \parallel$, the greater the ability to correct errors, but also the slower the communication because more characters must pass through the channel. 

Shannon's entropy \cite{shannon-weaver-49}, formally similar to thermodynamic entropy, is maximum when characters are equiprobable. It is an average of the  information  obtained by receiving one character. Its rationale is that the more uncertain the receiver is about which character she will receive, the more information she obtains upon receiving it.

In ET, testimonies are transmitted to a judge  for  evaluation. Thus, the context is that of a communication channel   \cite{shafer-tversky-85CS}. This is particularly evident in the 1-dimensional example mentioned in \S~\ref{sec:ET}, where the numerator of eq~\ref{eq:dempster-shafer} was explained in terms of serial testimonies whereas the denominator reflected parallel testimonies.

\begin{figure}
\center
\fbox{\resizebox*{0.8\textwidth}{!}{\includegraphics{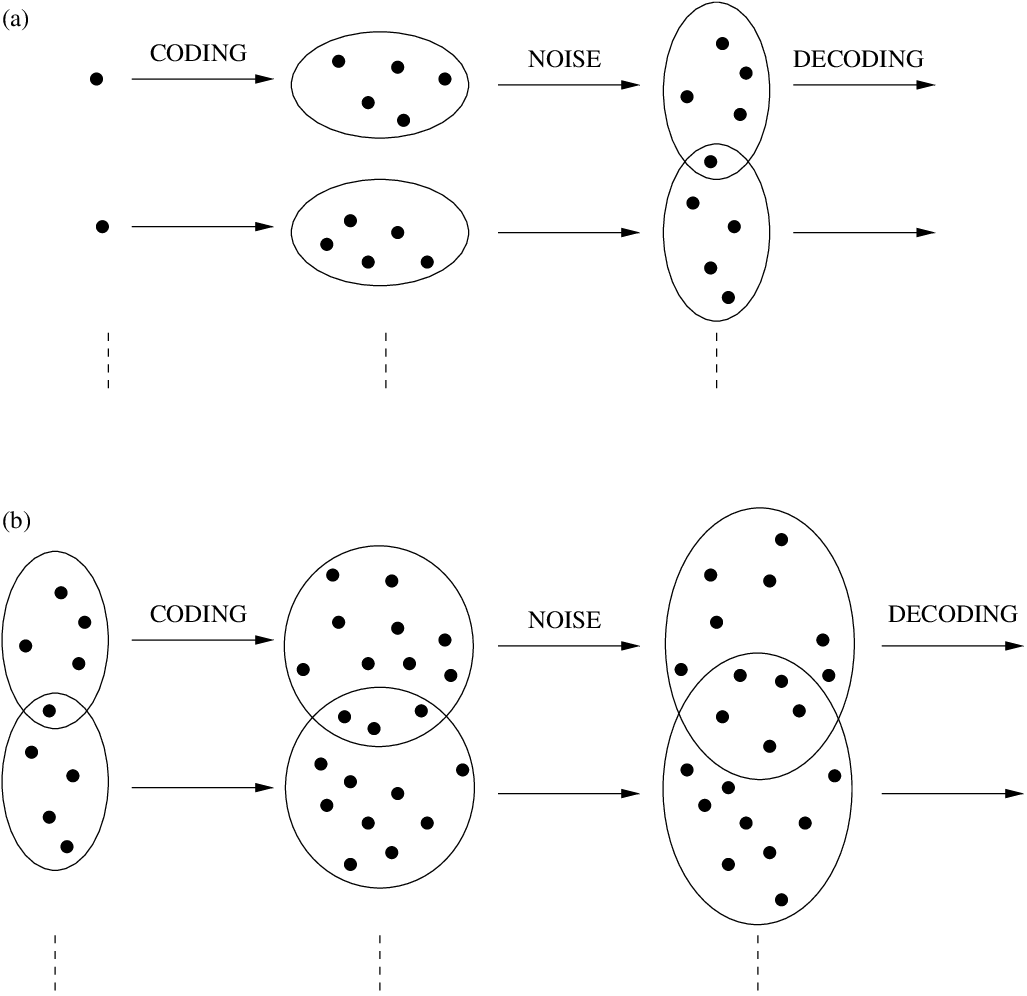}}}
\caption{In (a), the classical IT framework  where single characters $\{A_i\}$ are first coded into sets $A_i$, then transmitted through a noisy channel which may generate intersections between these sets, and finally decoded. In (b), the fusion of partially overlapping information originating from different sources. The original overlap may be enhanced by coding and further enhanced by transmission through a noisy channel.} \label{fig:fusion}
\end{figure}

In data fusion, ET generalizes the IT framework to multiple sources emitting partially overlapping information whose overlap is further enhanced by coding and subsequently by transmission through a noisy channel. Figure~\ref{fig:fusion} illustrates this generalization.

With multiple sources of partially overlapping sets, the definition of information entropy must be modified. The quest for an entropy function suitable for ET is a very active research field that did not yet reach a universally accepted functional (see \cite{klir-06} \cite{abellan-17CSF} \cite{moralgarcia-abellan-21IJIS} \cite{dezert-tchamova-22IS} \cite{jirousek-kratochvil-shenoy-22IJAR} for discussions and reviews). The following recent proposal \cite{ramisetty-jabez-subhrakanta-23IS}   is indicative of the sort of expressions that are being discussed:

\begin{equation}
  H(A) \; = \; - \sum_{A_i \in \Theta} \; \frac{Pl(A_i) \;\; \lg \: Pl(A_i)}{e^{Pl(A_i) -Bel(A_i)}} \;\; + \sum_{A_i \in \Theta} \; \big[Pl(A_i) -Bel(A_i) \big]   \label{eq:panda}
\end{equation}
where eqs.~\ref{eq:belief} and~\ref{eq:plausibility} have been applied with $\mathcal{H} \equiv A_i$.

The first term of eq.~\ref{eq:panda}  reduces to Shannon's entropy if $A_i \equiv \{A_i\}$, which implies that $Bel(A_i) =$ $ Pl(A_i) =$ $ p(\{A_i\})$. In the context of telecommunications, this term is maximum if characters are equiprobable. In ths context of ET, this term is maximum if alternative possibilities are supported by equal evidence.

In the closed world of telecommunication, the first term can only be minimized by adopting redundant codes that allow receivers to (partially) correct the mistakes introduced by the noisy channel (portion (a) of  Figure~\ref{fig:fusion}). However, ET applies also to open worlds such as communication between living organisms.

Living beings can give novel meanings to novel  possibilities generated by  random mutations. Whenever this happens, novel possibilities are added to the FoD and $H(A)$ can decrease because the number of possibilities has increased \cite{atlan-74JTB} \cite{atlan-87PS}.

The second term of eq.~\ref{eq:panda} has no counterpart in Shannon's entropy. The difference between $Pl(A_i)$ and $Bel(A_i)$ measures to what extent the available evidence goes beyond $A_i$  to support some other possibility as well (portion (b) of  Figure~\ref{fig:fusion}); thus, this term measures the ambiguity of communication codes. This is particularly important for human communication, where misunderstandings can generate novel possibilities.

\section{Decision Making by Seeking Coherence}           \label{sec:CSN}

Albeit utility maximization is the most widely employed model of decision-making,  experiments on preference reversal demonstrate that humans do not evaluate utility and probability independently of one another \cite{lichtenstein-slovic-06tutto} \cite{kim-seligman-kable-12FN} \cite{chen-gao-mcfadden-20FQP}. This is not a mere bias signalling that the basic model requires corrections and adaptations, but rather an indication that a decision model based on these two magnitudes can not reflect reality. Apparently, humans do not make their decision by evaluating two magnitudes, but just one.

In ET, either the belief or plausibility expressed by eqs.~\ref{eq:belief} and~\ref{eq:plausibility} are meant to express this magnitude \cite{shafer-86SS} \cite{shafer-16IJAR}. In ET, a decision is made as soon as Sherlock Holmes has tightened and coarsened the FoD until arriving at a  coherent interpretation of what looked like messy information. Decision is made by seeking coherence.

Note, incidentally, that understanding human decision as seeking coherence blurs the difference between individual and collective decisions. For instance, Dr. Watson may come up with details that stimulate Sherlock Holmes, and the final decision is made when Watson agrees with Sherlock Holmes.

Henceforth, I shall review one basic model of coherence-based decision that has been developed independently of ET, two models that include elements of ET, and finally outline the requirements of a model that should include the most advanced features of ET. The basic model is Constraint Satisfaction Networks (CSNs).

CSNs are neural networks whose neurons may represent possibilities, or concepts, or propositions linked to one another by either excitatory or inhibitory connections that represent inferences. Thus, an excitatory connection from neuron A to neuron B means ``A implies B'' whereas an inhibitory connection  means ``A implies $\neg$ B.''

The output of a neuron is the higher, the more and the stronger its excitatory inputs; conversely, inhibitory inputs tend to decrease output. The connection between any two nodes $i$ and $j$ is weighted by a coefficient $w_{ij}$ which is increased at each time step depending to its contribution to neuron output (Hebbian Rule). Updating is reflexive, with $\Delta \, w_{ij} = \Delta \, w_{ji}$. One notable property is that feedbacks between neurons make the network maximize  $Consonance =  \sum_{i,j} w_{ij}  y_i y_j$.

Consonance maximization means that those neurons  are strengthened, that represent possibilities, concepts or propositions which are coherent with one another. Thus, CNSs can be used to model decision-making as a search for coherence \cite{thagard-00}. Notable applications of CSNs are the elaboration of scientific theories by arranging empirical findings in a network of coherent causal relations,  the evaluation of guilt or innocence in a trial by fitting testimonies in a coherent frame, as well as the formation of medical diagnoses out of disparate analyses and symptoms   \cite{thagard-92} \cite{holyoak-simon-99JEPG} \cite{thagard-99} \cite{lundberg-04EJOR} \cite{lundberg-07EJOR}.

Recently, CSNs have been extended into networks of networks, where the inner networks represent concepts in individuals' minds \cite{rodriguez-bollen-ahn-16PLOS-ONE} \cite{bhatia-golman-19JMP} \cite{dalege-galesic-olsson-24PR} \cite{chopard-raynaud-stalhandske-25E}. Many among these models differentiate themselves  substantially from the original CSNs.

Evidential Networks (ENs) apply Evidence Theory to situations where all evidence is consonant, i.e., $\forall i$ it is $A_i \subseteq A_{i+1} \subseteq$ $A_{i+2} \ldots$. These networks have a tree structure (a directed acyclic graph). Since there are no partial intersections between possibilities, eq.~\ref{eq:dempster-shafer} reduces to a straightforward extension of Bayes's Theorem \cite{smets-93IJAR} \cite{xu-smets-96IJAR}. ENs are able to express missing information by means of $m(\Theta)>0$. In general, applications make hypotheses $\mathcal{H}$ coincide with distinct possibilities that include all others, hence no human intervention is required \cite{friedberg-hong-mclaughlin-smith-miller-17IEEE-A}.

Valuation Networks (VNs) exploit the  potential of ET in terms of intersecting possibilities \cite{shafer-shenoy-mellouli-87IJAR}  \cite{shafer-shenoy-90XIV} \cite{shenoy-00EJOR} \cite{benyaghlane-smets-mellouli-03XXIV}. VNs can be represented as  a hypergraph whose hyperedges  correspond to the possibilities envisaged in the FoD. Intersections between possibilities correspond to common faces between hyperedges; for instance, possibilities $A_i = \{\alpha, \beta, \gamma\}$ and $A_j = \{\beta, \gamma, \delta\}$ are triangles that have in common the edge $C_k = \{\beta, \gamma\}$.

\begin{figure}
\center
\fbox{\resizebox*{0.8\textwidth}{!}{\includegraphics{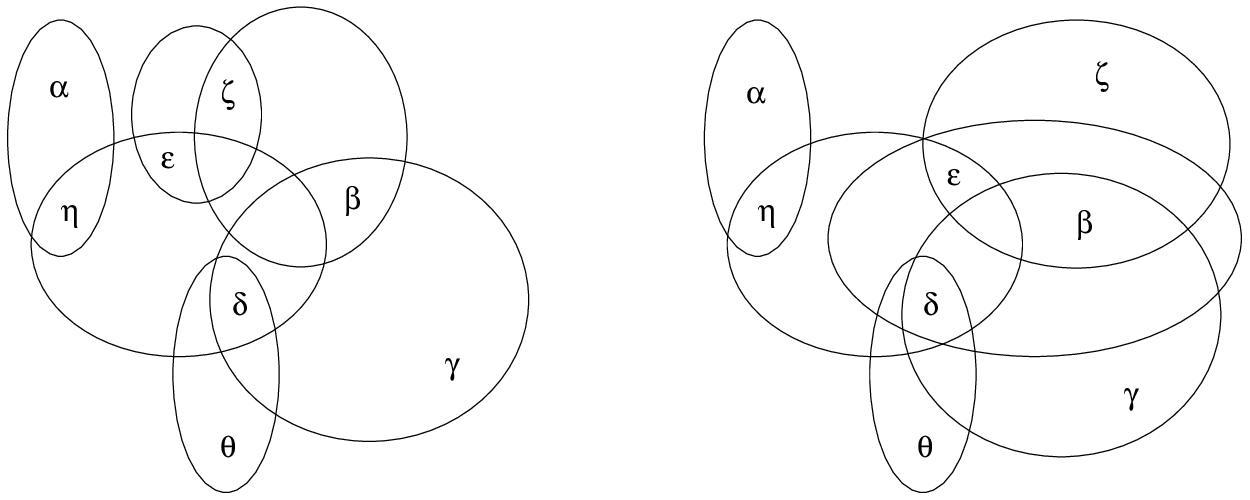}}}
\caption{The possibility sets corresponding to a cyclic hypergraph (on the left) and an acyclic hypergraph (on the right). For each hypergraph, the vertices of hyperedges are the elements entailed by the possibility sets, denoted by Greek letters. For instance, the possibility set  $A_1 = \{\alpha, \eta\}$ corresponds to a hyperedge (a segment) of vertices $\alpha$ and $\eta$. The cyclic hypergraph on the left is made of hyperedges  $A_1 = \{\alpha, \eta\}$, $A_2 = \{\beta, \zeta\}$, $A_3 = \{\delta, \theta\}$, $A_4 = \{\epsilon, \zeta\}$, $A_5 = \{\beta, \gamma, \delta\}$, $A_6 = \{\delta, \epsilon, \eta\}$. The acyclic hypergraph on the right can be obtained by removing $A_2$ and $A_4$ and adding $A_6 = \{\beta, \delta, \epsilon\}$ and $A_7 = \{\beta, \epsilon, \zeta\}$. Thus, the acyclic hypergraph has been obtained by coarsening the FoD. Freely redrawn from  \cite{shafer-shenoy-90XIV}.} \label{fig:hypergraphs}
\end{figure}

While ENs are directed acyclic graphs (trees), VNs are directed acyclic hypergraphs (hypertrees). Figure~\ref{fig:hypergraphs} illustrates the difference between the possibility sets that correspond to a cyclic hypergraph (on the left) and those that correspond to an acyclic hypergraph (on the right). In general, cyclic hypergraphs can be turned into acyclic hypergraphs by coarsening the frame of discernment. For instance, the acyclic hypergraph that corresponds to the sets on the right of Figure~\ref{fig:hypergraphs} can be derived from the cyclic hypergraph on the left by removing $A_2 = \{\beta, \zeta\}$ and $A_4 = \{\epsilon, \zeta\}$ and adding $A_6 = \{\beta, \delta, \epsilon\}$ and $A_7 = \{\beta, \epsilon, \zeta\}$ \cite{shafer-shenoy-90XIV}.

In VNs, the presence of a substantial $m(\Theta)>0$ may suggest FoD restructuration. In particular, coarsenings can often ease computation with little or no information waste \cite{benyaghlane-denoeux-mellouli-01XXXII} \cite{haduong-08IJAR}. Note that although any coarsening can be formulated, FoD refinement --- which  implies envisaging novel possibilities --- is not possible in in VNs.

Perspectively, let me label Open World Networks (OWNs) a future class of hypergraphs that either following $m(\Theta)>0$ or  $m(\emptyset)>0$ can implement both coarsenings and refinements of the FoD. In particular, novel possibilities entering the FoD must interact with the previous ones in ways that cannot be constrained along a pre-determined sequence. Thus, OWNs must be undirected cyclical hypergraphs.

OWNs should avoid that cycling information generates coherence independently of the soundness of the arguments that support it. Separating novel evidence from what has already been used could be a criterion to avoid unproductive cycling  \cite{gasparri-fiorini-dirocco-panzieri-IEEE-SMC-C}.

Table~\ref{tab:Networks} compares CSNs, ENs, VNs and OWNs with respect to (a) Node output, (b) Updating rules, (c) Structures, and (d) Objective functions.

\begin{table}    
\begin{center}
\begin{tabular}{|c||c|c|c|c|}  \hline
  & \multirow{2}{*}{(a)} & \multirow{2}{*}{(b)} & \multirow{2}{*}{(c)} & \multirow{2}{*}{(d)}  \\ 
  & \multirow{2}{*}{Output} & \multirow{2}{*}{Update} & \multirow{2}{*}{Structure} & \multirow{2}{*}{Objectives} \\ 
  &  \hspace{6em} &  \hspace{6em} &  \hspace{6em} &  \hspace{6em} \\ \hline\hline
 & weighted $\Sigma$  & \multirow{2}{*}{Hebbian} & Undirected &  \\ 
CSN & + excitatory & \multirow{2}{*}{Rule}  & Cyclic & $Consonance$ \\
 &  - inhibitory &  & Graph & \\  \hline
 & \multirow{2}{*}{$\mathcal{H}^A \equiv A_i \supseteq \forall A_j$}  & \multirow{2}{*}{Dempster} & Directed & \\ 
EN & \multirow{2}{*}{$\mathcal{H}^B \equiv B_h \supseteq \forall B_k$}  & \multirow{2}{*}{Shafer} & Acyclic & $m(\Theta)$ \\ 
 &   &  & Graph & \\  \hline
 & \multirow{2}{*}{$\forall \; \mathcal{H} \in 2^{\Theta}$}  & \multirow{2}{*}{Dempster}  & Directed & \\ 
VN & \multirow{2}{*}{coarsening $\Theta$} & \multirow{2}{*}{Shafer} & Acyclic &  $m(\Theta)$ \\ 
 &  &   & Hypergraph & \\  \hline
 & $\forall \; \mathcal{H} \in 2^{\Theta}$  &   & Undirected & \\ 
OWN & coarsening $\Theta$ & Smets & Cyclic &  $m(\emptyset)$, $m(\Theta)$ \\ 
 & tightening $\Theta$ &   & Hypergraph & \\  \hline
 \end{tabular}
\end{center}
\caption{Differences and similarities between CSNs, ENs, VNs  and OWNs with respect to (a) The output generated by single nodes, and the interactions it eventually stimulates in terms of formulating $\mathcal{H}$ or coarsening/tightening $\Theta$; (b) The rules for updating those outputs; (c) The structure of the network, and (d) Objective functions to be maximized or minimized ($2^{\Theta}$ denotes the set of all subsets of $\Theta$).} \label{tab:Networks}
\end{table}

All of these networks maximize or minimize an objective function. CSNs maximize \emph{Consonance}. ENs and VNs minimize  $m(\Theta)$. OWNs minimize both  $m(\Theta)$ and $m(\emptyset)$.

CSNs and OWNs have a similar structure (undirected cyclic graphs or hypergraphs), whereas ENs and VNs are structurally similar to one another (directed acyclic graphs or hypergraphs). ENs and VNs are similar to one another also insofar they both combine evidence by means of Dempster-Shafer's rule~\ref{eq:dempster-shafer}, whereas CSNs and OWNs are quite different from one another because CSNs employ the Hebbian rule whereas OWNs make use of Smets's rule~\ref{eq:smets}.

\section{Conclusions}  \label{sec:conclusions}

While ET is being increasingly used in data fusion,  its ability to deal with non-probabilistic uncertainty has been largely neglected. At the same time, non-probabilistic forms of uncertainty are being increasingly debated in the social sciences, pinning down definitions and differences but without any ability to develop mathematical and computational methods. By connecting these two research areas, I hope to  favour the awareness and usage of proper tools.

ET is fascinating because of  its unconventional assumptions. In particular, the lack of  complementation  somehow parallels the dismissal of classical, algorithm-based artificial intelligence as the true model for the human brain by the connectionist revolution.  It is  an uneasy choice, because instead of providing a ready-made algorithm that (supposedly) reproduces human uncertain reasoning, ET limits itself to offering suggestions for subsequent refinements or coarsening of the frame of discernment in the course of an interactive process that interrogates reality, formulates hypotheses, and back again. No ready-made solution, just a broad guidance in the quest for coherence.

Ascribing a positive mass to the void set is even more unconventional an assumption, for which careful mathematical foundations are in need. Intuitively, one may remark that  just like $m(\Theta)>0$ is not distributed among the $A_i$, the $m(\emptyset)>0$ is not distributed among anything. In a way, just like $m(\Theta)>0$ hovers above the FoD, the $m(\emptyset)>0$ hovers above the nihil.

Refusing to separate something like ``utility'' from something like ``probability'' is possibly the most striking feature of ET, one that runs against deeply ingrained ideas about what constitutes rationality. Thousands of years before utility maximization, \emph{The Fable of the Fox and the Winegrapes} pointed to the stupidity of the fox who, upon evaluating the probability to reach the grapes to be about zero, updated her utility by convincing herself that the grapes were sour. However, just like a Japanese \emph{koan}, this fable may hide deeper levels of understanding. One may notice, for instance, that the animal who was unable to separate utility from probability was not a donkey, but a fox. Perhaps, that animal was not so stupid.

\section*{Legal Disclaimer}

This research was not funded. The author has no conflict of interests. Neither humans nor other animals were involved in experiments.

\appendix

\section{Open Worlds}   \label{app:merge}

Assuming that novel events might appear, that are currently not in the frame of discernment, goes under the label of an \emph{open world}. By contrast, the expression \emph{closed world} means that all possibilities are known in advance. 

Shafer's seminal work on Evidence Theory  did not explicitly discuss this issue, albeit referencing judges or detectives rather than gamblers seems to imply an open world \cite{shafer-76}. Both the Transferable Beliefs Model and the Theory of Hints assume an open world which, after some aggregation, is reduced to a closed world regulated by probability at some higher level suitable for decision-making.

The assumption of an open world is closely tied to rejecting the assumption of a $\sigma-$algebra, which implies complementation and the definition of a residual event. Unfortunately, algebras are never mentioned in Shafer's seminal work \cite{shafer-76}. However, it is clear that the complementation operator is not available for incoming evidence, whereas it is available for the hypotheses that the judge or detective conceives. Moving from this observation, the Theory of Hints \cite{kohlas-monney-08XXVI} assumes that although the frame of discernment is not a  $\sigma-$algebra, hypotheses $\mathcal{H}$ are formulated within a subset $\Omega \subset \Theta$ which is a  $\sigma-$algebra. We followed this interpretation. By contrast, the Transferable Belief Model  defines the frame of discernment as a boolean algebra \cite{smets-92IJIS}. There are differences between  $\sigma-$algebras, which are employed in measure theory, and boolean algebras, which  are mainly employed in artificial intelligence. For our purposes, it suffices to say that both admit the complementation operator. Thus, with either a    $\sigma-$algebra or a boolean algebra the possibility set is exhaustive.

In spite of having assumed a  $\sigma-$algebra for the frame of discernment,  the Transferable Beliefs Model has been first to formalize the uncertainty that something may happen, that one is not even able to figure out by means of $m(\emptyset)>0$  \cite{smets-88} \cite{smets-92XL}. This is a clear departure from the basic Evidence Theory, where by definition $m(\emptyset)=0$ and $m(\Theta)=1$  \cite{shafer-76}. The rationale for this formalization is that  $m(\emptyset)>0$ when conflict between incoming evidence is very high, hence one suspects that something else may happen. The subsequent literature on open worlds has kept the convention and the rationale of the Transferable Beliefs Model \cite{smets-07IF} \cite{deng-15AI} \cite{daniel-16XV} \cite{yang-gan-tang-lei-20E}, which we did as well.

In the first version of Evidence Theory \cite{shafer-76}, \emph{weights of evidence} are empirical assessments $w(A_i):$ $ A_i \to [0, +\infty] \; \forall A_i$ that feed belief masses $m$ as follows:

\begin{equation*}
m(A_i) = 1- e^{w(A_i)}
\end{equation*}

Thus, with growing evidence the $m(.)$s grow while $m(\Theta)$ decreases. We interpreted $m(\Theta)$ as expressing uncertainty over known unknowns and, since we want to have a measure of uncertainty about unknown unknowns as well, we included both $m(\Theta)$ and  $m(\empty)$ in the normalization equation~\ref{eq:m}.

Evidence Theory \cite{shafer-76} called a testimony $m(A_1),$ $ m(A_2), \ldots$ a \emph{basic probability assignment}. This terminology induced much of the subsequent literature to understand it as an extension of Probability Theory. Indeed, most of the subsequent literature understood the frame of discernment as a $\sigma$-algebra. We rather followed the Transferable Beliefs Model in ignoring ``basic probability assignments'' altogether, assuming the existence of  masses $m(.)$ and defining belief and plausibility functions upon them \cite{smets-kennes-94AI}. According to this interpretation, Evidence Theory is built on empirical evidence, the frame of discernment on which it impinges, and the beliefs and hypotheses that humans entertain. Probability never enters the picture.

All versions of Evidence Theory are based on evidence combination rules, but some of them modify the original  eq.~\ref{eq:dempster-shafer}. For instance, the Transferable Beliefs Model eliminates the denominator that redistributes conflicting evidence. The appropriateness of Dempster-Shafer combination rule will be discussed at greater length in  \S~\ref{app:dezert}.

\section{Zadeh's Criticism}   \label{app:dezert}

Zadeh criticised Depster-Shafer's combination rule by means of the following example \cite{zadeh-84AIM}:

\begin{quotation}
    Suppose that a patient, P, is examined by two doctors, A and B. A's diagnosis is that P has either meningitis, with probability 0.99, or brain tumor, with probability 0.01. B agrees with A that the probability of brain tumor is 0.01, but believes that it is the probability of concussion rather than meningitis that is 0.99.
\end{quotation}

Application of   eq.~\ref{eq:dempster-shafer} leads to the conclusion that $Bel(\mathrm{Tumor}) = 1.0$, which is clearly unrealistic. Zadeh's paradox originates from mutually exclusive possibilities in a closed world, and it would generate equally paradoxical results   if Bayes' Theorem were applied without adding the possibility that either doctor is wrong \cite{smets-07IF} \cite{haenni-05WP}.

However, Zadeh's paradox  induced the belief that Dempster-Shafer's combination rule is wrong and sparked a search for alternatives. We grouped these and other alternatives  in three categories that are illustrated the ensuing \S~\ref{subsec:redistribute}, \S~\ref{subsec:channeling} and~\S~\ref{subsec:redistribute}, respectively.

\subsection{Redistributing Conflict}   \label{subsec:redistribute}

Since Zadeh's paradox derives from conflicting evidence, alternative combination rules have been devised that redistribute it among non-conflicting possibilities more efficiently. The list of these alternative rules is quite long \cite{sentz-ferson-02WP} \cite{zhu-li-04WP} \cite{ma-jiang-luo-19ASC}. 

Let us report the PCR5 rule, which is generally appreciated for the sensible results that it generates \cite{dezert-smarandache-15I} \cite{smarandache-dezert-tacnet-15IV}. Further improvements are able to accept several bodies of evidence at a time, as well as weighting by reliability and importance \cite{smarandache-dezert-15VII} \cite{liu-liu-dezert-cuzzolin-20IEEE-TFS}.

\begin{equation*}
    m_{PCR5}(C_k) \;\; = \; \sum_{A_i \cap B_j} \; m_A(A_i) \: m_B(B_j) \;\; + \; \sum_{X \cap C_k = \emptyset}\; \left[ \:  \frac{m_A^2(C_k)\: m_B(X)}{m_A(C_k)+ m_B(X)} \; \frac{m_B^2(C_k)\: m_A(X)}{m_B(C_k)+ m_A(X)} \;  \right]
\end{equation*}
where $X \in 2^\Theta$.

\subsection{Channelling Conflict Elsewhere}     \label{subsec:channeling}

Dempster-Shafer's combination rule (eq.~\ref{eq:dempster-shafer}) redistributes conflicting evidence among non-conflicting possibilities in ways that have been criticised ensuing Zadeh's criticism. The PCR5 rule and other rules mentioned in \S~\ref{subsec:redistribute} redistribute conflicting evidence among non-conflicting possibilities in other ways.

One radical alternative is that conflicting evidence is not redistributed among non-conflicting possibilities. Alternative proposals include transfering conflicting evidence to the frame of discernment $\Theta$, transfering it to the union of conflicting possibilities, as well as transfering it to $\emptyset$ \cite{smets-07IF}, which is the stance we have embraced.

Indeed, eq.~\ref{eq:smets} is alternative to eq.~\ref{eq:dempster-shafer}. If  eq.~\ref{eq:smets}  is applied to Zadeh's example, it delivers $m(\emptyset) = 0.9999$. That is, by listening to two experts saying opposite things one concludes that the truth might possibly lie somewhere else.

\subsection{Reframe the Problem}     \label{subsec:reframe}

An even more radical alternative builds on Shafer's own suggestion that the spirit of Evidence Theory is that the frame of discernment is refined and further refined until one finds pieces of evidence that are sufficiently detailed to be independent of one another \cite{shafer-16IJAR-hist}. In this specific case, the method consists of reframing the diagnoses of the two doctors, as well as the patient's understanding, in ways that are less crude than the coarse mutually exclusive possibilities envisaged by Zadeh.

For instance, Haenni remarked that diseases are not mutually exclusive, hence the doctor A rather expresses a probability 0.99 for meningitis either alone, or in conjunction with either tumor or concussion, or both. With a similar re-interpretation of doctor B's diagnosis, Dempster-Shafer's rule yields sensible outcomes \cite{haenni-05WP}. Likewise, Boivin solved the puzzle by reasoning that the patient  computes the union of the two diagnoses, obtaining similar conclusions \cite{boivin-22WP}.

This alternative does not actually contradict the solution of moving the conflicting evidence into an $m(\emptyset)>0$ mentioned in \S~\ref{subsec:channeling}. Indeed, in the course of the paper we have advocated both of them.

\section{From Dempster-Shafer to Bayes' Theorem}   \label{app:bayes}

In this appendix we show that, under assumptions~\ref{cond:radicalUncertainty} and~\ref{cond:radicalUncertainty}, Dempster-Shafer combination rule (eq.~\ref{eq:dempster-shafer}) reduces to Bayes' Theorem. Let us also assume that hypotheses are automatically generated by the combination of evidence through eq.~\ref{eq:dempster-shafer}, hence eqs.~\ref{eq:belief} and~\ref{eq:plausibility} are not necessary. Beliefs coincide with probabilities and, because of condition~\ref{cond:singletons}, beliefs at previous points in time can be expressed in terms of prior evidence  $\{ p(\{A_1\}),$  $ p(\{A_2\}), \ldots$ $ p(\{A_{N_A}\}) \}$.  Thus, updating beliefs means conditioning posterior probabilities on prior probabilities:

\begin{description}
  \item[Prior Probability:] $Bel^t(\mathcal{H}_i) \equiv p(\{A_i\})$, $\forall i$
  
  \item[Posterior Probability:]  $Bel^{t+1}(\mathcal{H}_i) \equiv p(\{A_i\} \mid \{B_j\})$, $\forall i,j$
\end{description}

For simplicity, and without loss of generality, let us assume that $\exists p, q: \{A_{p}\} \equiv \{B_{q}\}$ whereas  $ \forall i,j \neq \{p,q \}$ it is $ \{A_i\} \cap \{B_j\} = \emptyset$. Let us feed  eq.~\ref{eq:dempster-shafer} into $Bel^{t+1}(\mathcal{H}_p)$ while highlighting the time stamp by means of a superscript:

\begin{displaymath}
\begin{array}{ccccc}
  p(\{A_{p}^t\} \mid \{B_{q}^{t+1}\}) &
  \stackrel{(a)}{=} &
  \frac{p(\{A_{p}^t\}) \; p(\{B_{q}^{t+1}\})}{1 \; - \; \sum_{\substack{i\neq p \\ j \neq q}} \; p(\{A_i^t\}) \; p(\{B_j^{t+1}\})} &
  \stackrel{(b)}{=} &
  \frac{p(\{B_{q}^t\}) \; p(\{A_{p}^{t+1}\})}{p(\{B_{q}^{t+1}\})}  \\
  & & & & \\
                        &
  \stackrel{(c)}{=}     &
  \frac{p(\{B_{q}^{t}\} \mid \{A_{p}^{t+1}\}) \; p(\{A_{p}^{t+1}\})}{p(\{B_{q}^{t+1}\})} &
                        &
\end{array}
\end{displaymath}
which is Bayes' Theorem for $\{A_{p}\}$ and $\{B_{q}\}$.

Passage~(a) is a straightforward application of  eq.~\ref{eq:dempster-shafer}. The denominator of passage~(b) is obtained by remarking that albeit $\{A_{p}^t\}$ and $\{B_{q}^{t+1}\}$ overlap, $\{B_{q}^{t+1}\}$ comes at a later point in time. The numerator of passage~(b), as well as passage~(c), require a time inversion of the arrival of bodies of evidence  $\{ p(\{A_1\}), p(\{A_2\}), \ldots p(\{A_{N_A}\}) \}$ and  $\{ p(\{B_1\}), p(\{B_2\}), \ldots p(\{B_{N_B}\}) \}$, respectively. This is possible because the sequence of arrival  has no impact on Dempster-Shafer  rule and, in any case, it is the very same logic employed in the standard  demonstration of Bayes' theorem.

\bibliographystyle{plain}
\bibliography{references}
\end{document}